\documentclass{article}

\usepackage{arxiv}

\usepackage[utf8]{inputenc} 
\usepackage[T1]{fontenc}    
\usepackage{hyperref}       
\usepackage{url}            
\usepackage{booktabs}       
\usepackage{amsfonts}       
\usepackage{nicefrac}       
\usepackage{microtype}      
\usepackage{lipsum}
\usepackage{graphicx}
\usepackage{caption}
\usepackage{subcaption}
\graphicspath{ {./images/} }

\title{Gender prediction using limited Twitter Data}

\author{
 Maaike Burghoorn and Maaike de Boer and Stephan Raaijmakers\\
 Data Science, TNO\\
 The Hague, The Netherlands\\
 \texttt{maaike.burghoorn@tno.nl, maaike.deboer@tno.nl, stephan.raaijmakers@tno.nl} 
}

\begin{document}
\maketitle
\begin{abstract}
Transformer models have shown impressive performance on a variety of NLP tasks. Off-the-shelf, pre-trained models can be fine-tuned for specific NLP classification tasks,  reducing the need for large amounts of additional training data. However, little research has addressed how much data is required to accurately fine-tune such pre-trained transformer models, and how much data is needed for accurate prediction. This paper explores the usability of BERT (a Transformer model for word embedding) for gender prediction on social media. Forensic applications include detecting gender obfuscation, e.g. males posing as females in chat rooms. A Dutch BERT model is fine-tuned on different samples of a Dutch Twitter dataset labeled for gender, varying in the number of tweets used per person. The results show that finetuning BERT contributes to good gender classification performance (80\% F1) when finetuned on only 200 tweets per person. But when using just 20 tweets per person, the performance of our classifier deteriorates non-steeply (to 70\% F1). These results show that even with relatively small amounts of data, BERT can be fine-tuned to accurately help predict the gender of Twitter users, and, consequently, that it is possible to determine gender on the basis of just a low volume of tweets. This opens up an operational perspective on the swift detection of gender.

\end{abstract}


\section{Introduction}

In recent years, the field of Natural Language Processing (NLP) has experienced an important quality boost with Transformer models: attention-based deep learning models for word embeddings. One type of the state-of-the-art transformer models is named Bi-directional Encoder Representations from Transformers (BERT) \cite{devlin2018bert}. Compared to predecessors like word2vec \cite{mikolov2013efficient}, BERT has shown great performance over several domains, such as the biomedical domain \cite{lee2020biobert} and the scientific domain \cite{beltagy2019scibert}, and on different tasks such as named entity recognition, relation extraction and question answering. This performance boost comes with a cost: Transformer models typically have billions of parameters, are therefore slow to train and demand large amounts of data. Fine-tuning existing pre-trained BERT models partly solves this problem. Here, fine-tuning involves the adaptation of word embedding vectors for words on the basis of a loss function that measures the error on an additional labeling task, like topic classification (or any other NLP task assigning labels to text).

This paper addresses the following question: what is the trade-off between accuracy and supplementary data when fine-tuning BERT for an NLP classification task? We address this question in a security-oriented use case: the identification of gender in social media (Twitter) messages. This topic has an application for detecting e.g. pedophiles on social media: adults posing (grooming) as children, by adjusting their vocabulary and linguistic style. A lot of research has been done on gender prediction on social media data, such as tweets \cite{mukherjee2017gender}\cite{abdul2019sentence}. However, very little research has addressed how much data is exactly required to achieve state-of-the-art performance for these types of classification tasks. In this paper the usability of BERT for gender classification on social media, specifically Twitter, is explored. We  focus on the Dutch language and measure the performance of BERT when used for gender classification on a Dutch twitter corpus with varying number of tweets per person.

The next section provides related work about both BERT and gender classification. Section 3 contains the method with the description of the dataset, pre-processing and the specific BERT model used. Section 4 shows the results and section 5 concludes the paper.

\section{Related Work}

\subsection{Gender classification}
Mukherjee et al. \cite{mukherjee2017gender} have shown that classifying gender using standardized algorithms such as Naïve Bayes and maximum entropy classifiers can be improved by including features that capture the authorial writing style of social media users. The study achieved an accuracy of 71\%, which outperformed several commercially available gender prediction software tools that were available at the time. A similar study by Lopes et al. \cite{lopes2016gender} showed even better results for Portuguese tweets, testing a large variety of features extracted from the textual data. Using Best First decision trees they achieve an accuracy of 82\%. 

Other ways to profile social media users based on gender were also studied. In Alowidbi et al. \cite{alowibdi2013language} a language-independent gender classifier is trained on Twitter data using color features such as profile background color, text color and side-bar colors. The study shows that standardized algorithms such as Naïve Bayes and Decision trees can achieve F1-scores nearing 75\%.

Research on Dutch social media has also been conducted. In van der Loo et al. \cite{van2016text} both age and gender classification experiments were performed on a Dutch chat corpus with nearly 380.000 posts from a social network. Using several features such as token and character n-grams, they achieved a F1-score of around 69\% when predicting gender using a SVM classifier. Another Dutch study \cite{van2017profiling} aimed to profile social media users by predicting their political preference and income level (low vs high) using several combinations of word unigrams, bigrams and trigrams. They achieve an F1-score of 72\% with the best setup for income class prediction and an accuracy of 66\% with the best setup for political preference prediction.

Several datasets for the task of gender classification have been created, such the Arap-tweet dataset  \cite{zaghouani2018arap} and the Netlog dataset \cite{van2016text}. Other studies have generated a dataset as part of their research \cite{mukherjee2017gender} \cite{alowibdi2013language} \cite{lopes2016gender}. The corpus used in this analysis is the TwiSty corpus developed by Verhoeven et al. \cite{verhoeven2016twisty} for the research of author profiling. The TwiSty corpus includes Dutch tweet data with annotated gender labels and has a balanced male/female user distribution. Research by the same authors showed that the gender of the Twitter users in the TwiSty dataset can be predicted with an F1-score between 73\% and 88\% for different languages using several standardized machine learning algorithms, such as linear SVC and logistic regression. The results for the Dutch model near an F1-score of 83\%. However, to the best of our knowledge BERT has not been tested on TwiSty. Furthermore, an analysis on the number of tweets needed for a solid prediction on Dutch Twitter data is also not reported in literature yet.

\subsection{BERT}
\begin{figure}[htbp]
\includegraphics[scale=0.4]{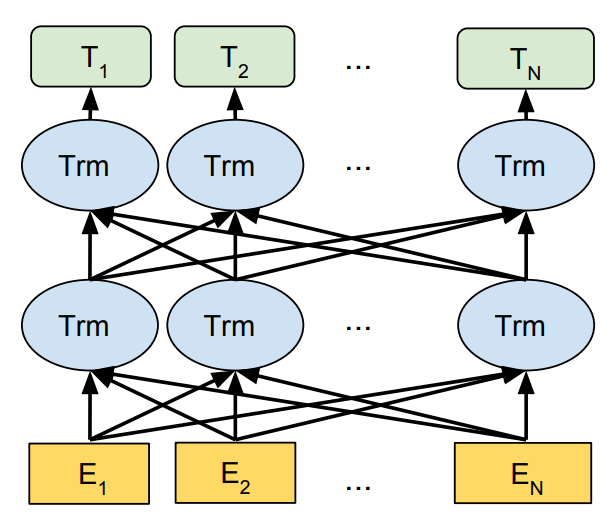}
\centering
\caption{Architecture of BERT, taken from \cite{devlin2018bert} }
\label{pic:1}
\end{figure}

BERT (Bi-directional Encoder Representations from Transformers) is a transformer model developed in 2019 by Devlin et al. \cite{devlin2018bert}. It was pre-trained on Wikipedia and Bookcorpus data and can be fine-tuned for specific tasks. The corpus on which BERT was trained, contains data for multiple languages, which makes BERT a multi-lingual model and allows it to be used for NLP tasks in multiple languages. As shown in Figure \ref{pic:1} BERT uses bidirectional transformers, which contrasts with traditional transformer models that only use left-to-right transformers or a concatenation of independently trained left-to-right and right-to-left transformers. In the transformer step BERT embeds a whole sequence of words at once. Before embedding, 15\% of the words are replaced with a mask token. The model then tries to predict the original values of these masked tokens based on the context provided by the other words in the sequence.

BERT has shown great performance in several different domains and on several different tasks. Unlike previous transformer models, BERT adds context to its word embeddings. This implies that words that have multiple meanings in different contexts are also assigned a different embedding in different contexts. A few examples of the tasks that BERT was used for are document retrieval, where BERT is use to create a ranking among documents based on a search query, \cite{yang2019simple}. Another task is conversational question answering, where a model is trained to perform information retrieval in an interactive manner, \cite{qu2019bert}, or sentiment classification where a model is used to classify the sentiment of digital media, \cite{gao2019target}. Some domains require BERT to be trained on domain-specific data. One of these domains is the biomedical domain in which a BERT model named BIOBERT \cite{lee2020biobert} was trained on a large-scale biomedical corpora. The paper shows that BIOBERT has great performance on three biomedical text mining tasks: named entity recognition, biomedical relation extraction and biomedical question answering. Another great application of BERT was shown as SciBERT \cite{beltagy2019scibert}, a pre-trained language model for scientific text. SciBERT was trained on a large corpus of scientific text and has shown impressive state-of-the-art (SOTA) performance on several domains, such as the biomedical domain and the computer science domain. Both the BIOBERT and SciBERT models are examples of models trained on large (text) corpora. The main problem with training these new BERT models is that this requires an extensive amount of data which is not always readily available. One application that uses fine-tuning of a BERT model is described in Adhikari et al. \cite{adhikari2019docbert}. A model called Doc-BERT is created by fine-tuning base BERT for the task of document classification. The paper showed that BERT can achieve great performance when used for the task of document classification, which had not been shown before. However, even the studies that simply fine-tune a off-the-shelf BERT model still use an extensive amount of data for this task.

BERT is also already used in gender classification. A study by Abdul et al. \cite{abdul2019sentence} used BERT to predict gender and age both individually and in a multi-task setting. They used a set of 100 users who posted at least 2000 tweets per person. They achieve an accuracy of around 65\% on gender prediction.

\section{Method}
\subsection{Dataset}

In our experiments, we use the TwiSty corpus \cite{verhoeven2016twisty}, a corpus of gender- and age-labeled tweets. The TwiSty corpus consist of a separate dataset for six different languages. Each dataset contains the following columns: \textit{confirmed \_tweet\_id\textit{}s}, \textit{gender (M/F)}, \textit{MBTI}, \textit{other\_tweet\_ids} and \textit{user\_id}. The creators of the corpus divided the tweet ids across two different columns: \textit{confirmed \_tweet\_id\textit{}s} and \textit{other\_tweet\_ids}. The \textit{confirmed \_tweet\_id\textit{}s} were verified to be in written in the language of the dataset. The \textit{other\_tweet\_ids} column contains the remaining unchecked tweets. The \textit{gender (M/F)} column is used as our output label. \textit{MBTI} is a personality metric and not used in this research. The column \textit{user\_id} is used to merge the tweets from one user. Table \ref{table:1} shows the number of users and total number of tweets for each dataset as originally reported in \cite{verhoeven2016twisty}.

Twitter’s privacy policy does not allow for sharing the content of tweets, therefore only ids could be stored and the actual textual content of the tweet had to be scraped. This requires a Twitter Developer account with accompanying access and authentication tokens. The scraping was performed in batches of 100, 3 batches per user. Users with less than 300 tweets were removed from the data. For the larger datasets only 1500 users were randomly selected and scraped. The number of tweets scraped per dataset can be found in Table \ref{table:2}.

\begin{table}[h!]
\parbox{.45\linewidth}{
\centering
\begin{tabular}{ |p{6em}||p{4em}|p{6em}|  }
 \hline
 Language & \#users & Total \#tweets \\
 \hline
 German	& 411 & 713.744\\
 Spanish & 10.772 & 13.493.445\\
 French & 1.405 & 1.995.865\\
 Italian & 490 & 658.332\\
 Dutch & 1.000 & 1.541.259\\
 Portuguese & 4.090 & 6.353.763\\
  \hline
 Total & 18.168 & 24.756.408\\
 \hline
\end{tabular}
\caption{TwiSty corpus data overview}
\label{table:1}
}
\hfill
\parbox{.45\linewidth}{
\centering
\begin{tabular}{ |p{6em}||p{4em}|p{6em}|  }
 \hline
 Language & \#users & Total \#tweets \\
 \hline
 German	& 289 & 57.800\\
 Spanish & 1197 & 239.400\\
 French & 825 & 165.000\\
 Italian & 297 & 59.400\\
 Dutch & 720 & 144.000\\
 Portuguese & 1002 & 200.400\\
  \hline
 Total & 4.330 & 866.000\\
 \hline
\end{tabular}
\caption{Scraped corpus data overview}
\label{table:2}
}
\end{table}

\vspace{-20pt}

\subsection{Pre-processing}
The analysis focused on the Dutch dataset, with a total of 720 users. For each user the tweets were concatenated and no additional features were created for the purpose of classification. However several pre-processing steps were performed. URL’s, digits and non-ascii characters, such as the '@' and '\#' characters which are commonly used in tweets, were removed. The tweets were also filtered for stopwords and lemmatization was performed. 

\subsection{BERT}
In this paper, the BERT classification model was implemented in Python using the Simple Transformer library \cite{simpletransformers}. This library wraps around the existing Transformers library by Hugging Face \cite{Wolf2019HuggingFacesTS} and allows for easy implementation of transformer models. The Simple Transformer library can be implemented in just a few steps and allows for easy tweaking of parameters such as number of epochs, training and evaluation batch size and early stopping conditions. When initializing a classification model using the library, many different pre-trained state-of-art models can be loaded, such as BERT, XLNET or GPT. We used a Dutch BERT model, referred to as BERTje, developed at the University of Groningen \cite{de2019bertje}. This model was pretrained on a large and diverse dataset of 2.4 billion tokens, not only based on Wikipedia data but also other sources such as book and web news.

\subsection{Experimental Setup}
We created four different gender classifiers, each trained on a different number of tweets per person. The datasets contained \textit{20}, \textit{50}, \textit{100} or \textit{200} concatenated tweets per user. The performance for the various classification models trained on the different datasets are compared. For each dataset a split 90/10 (train/evaluation data) was used. After several test runs, it was found that the model overfits after several epochs therefore early stopping conditions were used. The maximum number of epochs was set to 20 and every epoch the classifier was evaluated using the MCC metric on a separate evaluation dataset. If the classifier did not improve its score with 0.01 for more than 5 consecutive evaluations, early stopping is triggered. The data was trained and evaluated in batches of 16 and the learning rate was set to Simple Transformers' default value $4*10^{-5}$. The other hyperparameters were also kept at the default values of the Simple Transformer's package \cite{simpletransformers}. Cross-validation was used to check for consistency of the results. The dataset was split in 10 separate folds. Each run one of the folds was used for evaluation and the remainder of data was used for training. After early stopping conditions were triggered, the best model was selected and the accuracy and F1 scores were calculated. The class distribution was checked for choice of metric.

\section{Results}
\begin{figure}[htbp]
\centering
\includegraphics[scale=0.5]{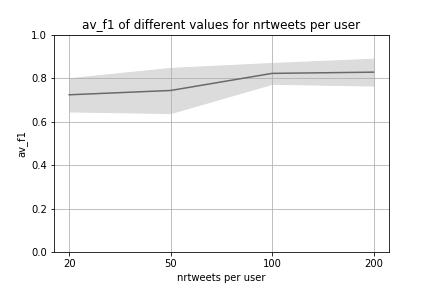}
\caption{Average F1-scores}
\label{pic:avf1}
\end{figure}

 The average F1-score over the 10-fold cross-validation was calculated and is shown in Table \ref{table:3}. A visualization of the results is shown in  Figure \ref{pic:avf1}.

\begin{table}[h!]
\centering
\begin{tabular}{ |p{7em}||p{4em}|p{4em}|  }
 \hline
 \multicolumn{3}{|c|}{Results} \\
 \hline
 nrtweetsperuser & av f1 & std f1\\
 \hline
 20  &	0.72 &	0.08\\
 50 &	0.74 &	0.10\\
 100 &	0.82	&	0.05\\
 200 &	0.83 &	0.06\\
 \hline
\end{tabular}
\caption{Results 10-fold cross validation}
\label{table:3}
\end{table}

The results show that when a concatenation of 200 tweets per person is used, a F1-score of 0.83 is reached. When the model is trained on only half the tweets per person, the score barely deteriorates to 0.82. Using even fewer tweets drops the F1-score to 0.74 for 50 tweets per user and 0.72 for 20 tweets per user. For each of the trained models, the standard deviation is relatively small, which shows the consistency of the results. The difference between the scores of the model using the most tweets (200) and the model using the least tweets (20) shows a non-steep decrease of only 0.1. For each of the models, the MCC score and the accuracy were also calculated, which show a similar trend in scores as the F1-score.

\section{Conclusion, Discussion and Future Work}
In this paper, the usability of BERT for gender prediction on social media data was explored. In particular, the amount of data that is required for accurate prediction was studied. A Dutch BERT model was fine-tuned on different samples of a Dutch Twitter dataset labeld for gender and varying in the number of tweets used per person. As shown in the results, the gender of Dutch twitter users can be accurately predicted using BERTje. Fine-tuning BERT on a larger dataset of 200 tweets per person achieves an impressive F1-score of nearly 83\%. These results are similar to results achieved by previous research, which confirms that with an extensive amount of data both traditional and transformer models can achieve great performance on author profiling classification tasks. In this paper it is also shown that when fine-tuning BERT with as little as 20 tweets per person the F1-score deteriorates non-steeply to 70\%. This is still a relatively high score for gender classification, also when compared to previous research, such as Abdul et al. \cite{abdul2019sentence} were an accuracy of 65\% was reached on a similar task. The improvement in results may be due to the different language on which the models were trained. This paper also showed that BERT can achieve better results on the gender prediction task than standardized machine learning algorithms, which were shown to have an F1 score in the range of 73\% to 88\% but with the trade-off of using more data and additional features \cite{verhoeven2016twisty}. This could be explained due to BERT being a pre-trained model and thus requiring less data when fine-tuning on a classification task.

The current results show that is it possible to determine gender on the basis of just a low volume of tweets. This is a promising result and opens up possibilities for near real-time detection of gender on social media. Future research can be performed to the study the effect of adding additional features to the BERT model, and use multi-task learning to not only predict the gender but also the personality of the author. BERT models are also known to produce interpretable attention patterns that contribute to the explainability of results \cite{clark2019does}. In our future work, we aim to analyze these patterns and make them available to potential end-users of this technology.

\paragraph{Acknowledgements}
We would like to acknowledge the H2020 ASGARD project for their financial support.

\bibliographystyle{unsrt}  
\bibliography{template}

\end{document}